%% file: main.tex
\documentclass[preprint,journal]{vgtc}            % preprint (journal style)

\onlineid{1066}

\preprinttext{To appear in IEEE Transactions on Visualization and Computer Graphics.}

\vgtccategory{Research}

\vgtcpapertype{Analytics \& Decisions}

\title{PromptMagician: Interactive Prompt Engineering for\texorpdfstring{\\}{}Text-to-Image Creation}

\author{%
    Yingchaojie Feng,
    Xingbo Wang,
    Kam Kwai Wong,
    Sijia Wang,
    Yuhong Lu,
    Minfeng Zhu,\texorpdfstring{\\}{}
    Baicheng Wang, and
    Wei Chen
}

\authorfooter{
  \item Y. Feng, S. Wang, Y. Lu, B. Wang, and W. Chen are with the State Key Lab of CAD\&CG, Zhejiang University. W. Chen is also with the Laboratory of Art and Archaeology Image (Zhejiang University), Ministry of Education. W.~Chen is the corresponding author. Email: \{fycj, sijiawang, luyuhong, wangbaicheng, chenvis\}@zju.edu.cn.
  \item X. Wang and KK Wong are with the Hong Kong University of Science and Technology. Email: \{xwangeg, kkwongar\}@connect.ust.hk.
  \item M. Zhu is with Zhejiang University. Email: minfeng\_zhu@zju.edu.cn.
}

\input{chapters/0-abstract}


\teaser{
  \centering
  \includegraphics[width=\linewidth]{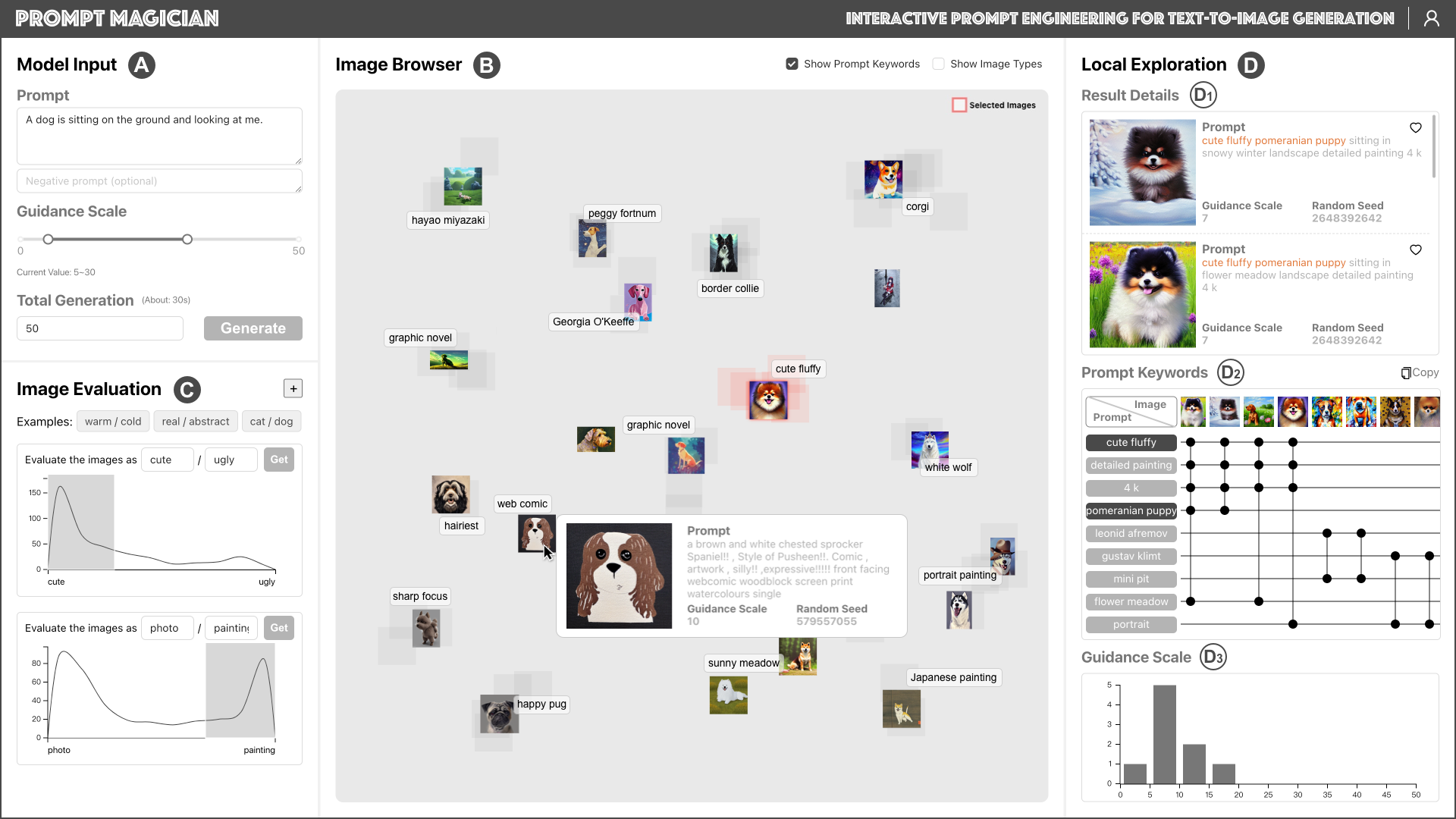}
  \caption{The user interface of \textit{PromptMagician} consists of four views. The \textit{Model Input View} (A) configures the prompts and hyper-parameters for image creation. The \textit{Image Browser View} (B) visualizes the generated and retrieved images and the recommended prompt keywords. The \textit{Image Evaluation View} (C) helps evaluate and filter images based on multiple criteria. The \textit{Local Exploration View} (D) helps users explore and validate the prompt keywords and guidance scales for images of interest.
  }
  \label{fig:teaser}
}

\graphicspath{{figs/}{figures/}{pictures/}{images/}{./}} % where to search for the images

\usepackage{tabu}                      % only used for the table example
\usepackage{booktabs}                  % only used for the table example
\usepackage{lipsum}                    % used to generate placeholder text
\usepackage{mwe}                       % used to generate placeholder figures

\usepackage{mathptmx}                  % use matching math 
\usepackage{amsmath}
\usepackage{comment}

\newcommand{\name}{{\textit{PromptMagician}}}

\usepackage{xspace,xpunctuate}

\newcommand{\ie}{\textit{i.e.},\xspace}
\newcommand{\etal}{\xspace\textit{et al.}\xspace}
\newcommand{\eg}{\textit{e.g.},\xspace}

\newcommand{\xingbo}[1]{\textcolor{black}{#1}}
\newcommand{\feng}[1]{\textcolor{black}{#1}}
\newcommand{\rc}[1]{\textcolor{black}{#1}}
\newcommand{\rw}[1]{\textcolor{black}{#1}}

\begin{document}

%%%%%%%%%%%%%%%%%%%%%%%%%%%%%%%%%%%%%%%%%%%%%%%%%%%%%%%%%%%%%%%%
%%%%%%%%%%%%%%%%%%%%%% START OF THE PAPER %%%%%%%%%%%%%%%%%%%%%%
%%%%%%%%%%%%%%%%%%%%%%%%%%%%%%%%%%%%%%%%%%%%%%%%%%%%%%%%%%%%%%%%

\maketitle

\input{chapters/1-introduction}
\input{chapters/2-related-work}
\input{chapters/3-design-process}
\input{chapters/4-method}
\input{chapters/5-user-interface}
\input{chapters/6-case-study}
\input{chapters/7-evaluation}
\input{chapters/8-discussion}
\input{chapters/9-conclusion}

%% if specified like this the section will be ommitted in review mode
\acknowledgments{
The authors would like to thank the anonymous reviewers for their insightful comments.
This paper is supported by the National Natural Science Foundation of China (62132017) and the Fundamental Research Funds for the Central Universities (226-2022-00235).
}

\bibliographystyle{src/abbrv-doi-hyperref}

\bibliography{main}

%% ^^^^^   FOR IEEE VIS, EVERYTHING HERE MAY BE INCLUDED IN THE    ^^^^^ %%
%% 2-PAGE ALLOTMENT FOR REFERENCES, FIGURE CREDITS, AND ACKNOWLEDGEMENTS %%

\appendix % You can use the `hideappendix` class option to skip everything after \appendix

\end{document}

%% file: chapters/0-abstract.tex
\abstract{
Generative text-to-image models have gained great popularity among the public for their powerful capability to generate high-quality images based on natural language prompts. However, developing effective prompts for desired images can be challenging due to the complexity and ambiguity of natural language. This research proposes {\name}, a visual analysis system that helps users explore the image results and refine the input prompts. The backbone of our system is a prompt recommendation model that takes user prompts as input, retrieves similar prompt-image pairs from DiffusionDB, and identifies special \feng{(important and relevant)} prompt keywords. To facilitate interactive prompt refinement, {\name} introduces \feng{a multi-level visualization for} the cross-modal embedding of the retrieved images and recommended keywords, and supports users in specifying multiple criteria for personalized exploration. Two usage scenarios, a user study, and \feng{expert interviews} demonstrate the effectiveness and usability of our system, suggesting it facilitates prompt engineering and improves the creativity support of the generative text-to-image model.
}

%% file: chapters/1-introduction.tex
\section{Introduction}
% The background of text-to-image and stable diffusion
Generative text-to-image framework has become a popular and effective interactive paradigm \cite{scao2021many} with widespread adoption in academia~\cite{ZhuP0019, nichol2021glide, ramesh2022hierarchical, rombach2022high} and the public~\cite{wang2022diffusiondb, liu2022design}.
The endless space of natural language text allows for the free expression of artistic ideas and significantly lowers the barrier to image creation.
With the rapid development of natural language processing (NLP) and computer vision (CV) technologies, state-of-the-art generative models, such as Stable Diffusion \cite{rombach2022high} and DALL$\cdot$E 2 \cite{ramesh2022hierarchical}, have been able to generate relevant and high-quality images based on text prompts and have demonstrated great potential in downstream tasks, including hyper-realistic video generation~\cite{abs-2210-02303} and radiology image synthesis~\cite{abs-2210-04133}.

% The introduction of prompt engineering
Building upon the success of these generative models,
researchers and developers have explored a human-model interaction technique called ``prompting''~\cite{gao2020making, schick2020s, weng2023towards}. 
During the creation process, users craft natural language prompts that describe the expected image characteristics (\eg subjects and styles), \feng{adjust model hyper-parameters (\eg guidance scales) and try out more seeds} to obtain the desired output.
% natural language complexity
However, the complexity and ambiguity of natural language can make it challenging for users, especially novice users, to develop effective prompts that trigger the model to generate the desired output~\cite{wang2022interactive}.
% \TODO{And this often results in a tedious, trial-and-error process of iterative prompt refinement.}
% model parameter space complexity
Additionally, prompts can result in distinct images based on different model hyper-parameters. 
It is difficult to evaluate the quality of the prompts with the limited trial of hyper-parameter values.
When receiving undesirable image results, users may become confused about whether and how to adjust the prompts or model hyper-parameters.
Previous research has proposed automatic prompting techniques \cite{wang2023reprompt} for text-to-image creation.
However, the image creation process greatly depends on human subjective judgment, which requires humans in the loop to refine the generation.
Some research \feng{\cite{wang2022diffusiondb, oppenlaender2022taxonomy}} suggests using ``magical spells'' (\eg keywords) to formulate prompts based on large human-annotated corpora. However, these guidelines could be too general to satisfy personalized image creation needs. 

% Our work
%     A workflow that guide users 
%     A visual interactive system
%     Cases and evaluation
% In this paper, we explore how to facilitate users in prompt engineering for text-to-image generation.

To address these challenges, we present {\name}\footnote{The code is available at \url{https://github.com/YingchaojieFeng/PromptMagician}}, a novel visual analysis system for interactive prompt engineering.
It aims to help users efficiently explore and evaluate the model-generated images and refine the input prompts and hyper-parameters for the desired image results.
Given a text prompt, the system automatically generates a collection of image results with a range of hyper-parameter values and retrieves related prompt-image pairs from DiffusionDB, a large prompt-image corpus \cite{wang2022diffusiondb}.
Then, the system presents a visual summary of both the generated and retrieved images to guide the exploration of images with diverse styles.
Additionally, users can specify image evaluation criteria using descriptive words (\eg ``good'' for image quality and ``beautiful'' for abstract perception) to filter out irrelevant images and focus on the image subset of interests for efficient exploration.

To assist users in prompt improvement, we propose a prompt keyword recommendation model based on prompt engineering design guidelines~\cite{liu2022design}, which prioritizes prompt keywords over sentence structures.
The model encodes the retrieved images from DiffusionDB with the CLIP model \cite{radford2021learning}, organizes them into hierarchical clusters, and identifies special \feng{(important and relevant)} prompt keywords from the corresponding prompts of the image clusters.
The importance of the keywords is measured using cluster-level TF-IDF values~\feng{\cite{sparck1972statistical}}.
Finally, the model matches the keywords with their most related clusters and recommends them to users for prompt improvement.
The recommended prompt keywords are visualized alongside the matched image clusters to facilitate user exploration.
The users can select image subsets to explore their prompt keywords and guidance scale.
% The system also helps users validate the effect of the recommended prompt keywords using the retrieved images.
\feng{The system also visualizes the interrelationship between the prompt keywords and images to help users understand and compare the effects of different prompt keywords for image creation.}
We evaluate our system and prompt recommendation model through two usage scenarios, a user study, \feng{and expert interviews}, and the results show that our system can help users discover effective prompt keywords and inspire image creation.

In summary, our major contributions include:
% \begin{itemize}[noitemsep, topsep=0pt]
\begin{itemize}
    \item A visual system to help users explore and evaluate the model-generated results and conduct interactive prompt engineering.
    \item A prompt recommendation model that identifies important and relevant prompt keywords to help prompt improvements.
    \item Two usage scenarios, a user study, and \feng{expert interviews} that demonstrate the effectiveness and usability of our system.
\end{itemize}

%% file: chapters/2-related-work.tex
\section{Related Work}
\subsection{Prompt Engineering}
With the rapid development of large language models \cite{ouyang2022training, brown2020language} and text-to-image models \cite{nichol2021glide, ramesh2022hierarchical, rombach2022high}, prompt engineering \cite{gao2020making, schick2020s, weng2023towards} has become a promising paradigm \feng{for interacting} with models \cite{lester2021power}.
With this paradigm, users can focus on designing and refining the prompt input to improve the performance of the pre-trained model in specific application scenarios, directly utilizing the knowledge and capability of the pre-trained model without the additional training process.
Nowadays, it has gained widespread attention and shown great potential in various tasks, such as natural language understanding \cite{jiang2020can}, image generation \cite{liu2022design}, and logical reasoning \cite{wu2022promptchainer}.

% Auto-prompt
Previous studies have focused on automatic approaches for prompt formulation and refinement.
AutoPrompt \cite{shin2020autoprompt} applied gradient-guided search in the collection of trigger tokens to automatically create prompts for masked language models.
Gao\etal \cite{gao2020making} employed the generative T5 model to generate the prompt templates and pruned brute-force search for label word selection.
To facilitate human-AI collaboration in prompt engineering \cite{wu2022ai},  interactive and visual systems were proposed.
% PromptChainer, PromptIDE
PromptIDE \cite{strobelt2022interactive} provides interactive visualizations to help users evaluate the performance of prompts on a small dataset and iteratively refine prompts.
For complex tasks that require multi-step operations, PromptChainer \cite{wu2022promptchainer} allows users to interactively construct chains of prompts for the corresponding targeted sub-tasks, increasing the transparency and controllability of large language models.

Most of the aforementioned studies are designed for text-to-text generative models whose output can be transformed into label results and used for the quantitative evaluation of prompt performance on a given dataset.
Our work focuses on text-to-image generative models, which have different outputs and evaluations \cite{wang2023reprompt}.
To provide guidelines for prompting research, Liu\etal \cite{liu2022design} conducted experiments to explore a set of open questions in prompt engineering for text-to-image models.
The results emphasized the importance of the prompt keywords (\ie subject and style) over the phrasing structures.
Based on an ethnographic study with community practitioners, Oppenlaender \cite{oppenlaender2022prompt} summarized a taxonomy of prompt modifiers, including subject terms, modifiers, and magic terms, to guide and inspire prompt formulation.
Nevertheless, these guidelines could be too general to satisfy personalized image creation needs.
A recent work, RePrompt \cite{wang2023reprompt}, introduces explainable AI techniques (\eg SHAP value \cite{lundberg2017unified}) to reveal the importance of text features, including the numbers and concreteness of each POS (part of speech) type, and their optimal value ranges.
Opal \cite{liu2022opal} utilized GPT-3 \cite{brown2020language} to generate text prompts for new illustrations.
Our work differs from prior work by combining database retrieval and ad-hoc generation, enabling users to explore the vast artistic search space to identify effective prompt keywords for personalized creation and iteratively refine the prompts.

\subsection{Visual Exploration of Image Collections}
A large number of daily-created images provide rich information for various applications, such as AI model development~\cite{chen2021towards} and content retrieval~\cite{xie2018semantic}.
Prior studies have proposed many techniques for image exploration at scale, such as tree-based visualizations~\cite{bertucci2022dendromap}, enhanced scatter plots~\cite{xia2022interactive, zhu2021visualizing, wang2021m2lens, wong_anchorage_2023}, and node-link graphs~\cite{zeng2019emoco,zeng2022gesturelens,liang2022multiviz}.

One major challenge is the summary and exploration of the complex semantics associated with images~\cite{wong_scrolltimes_2023, he2023videopro}.
Semantic Image Browser~\cite{yang2006semantic} annotates semantic content in images and uses a Multi-Dimensional-Scaling-based image layout that aggregates semantically similar images together. 
Similarly, Xie\etal~\cite{xie2018semantic} produced semantic image descriptions using image captioning techniques. 
Then, they utilized a co-embedding model to project images and their semantic descriptions into the same 2D space and employed a galaxy metaphor to provide a semantic overview of image collections.
Chen\etal~\cite{chen2021towards} proposed a node-link-based visualization powered by a co-clustering algorithm to reveal object labels and images in object detection tasks. The interactive visualizations help users explore and validate labels of the detected image objects.
Compared to prior work that mostly concentrates on image content exploration, we further consider other factors, such as image styles and model hyper-parameters, to help users formulate and refine their prompts to create visually appealing images using generative models.
For example, we use a pre-trained vision-language model, CLIP, to encode images, which considers both image content and visual styles. Based on model encodings, our system allows users to evaluate and filter images using natural language descriptions about image properties, such as ``cartoon'' and ``beautiful.''

\subsection{Text-to-Image Generation}
Text-to-image generation refers to translating natural language descriptions (\eg words and sentences) into realistic images.
Recent breakthroughs in computer vision (CV) and natural language processing (NLP) techniques have greatly improved text-to-image generation quality. 
Modern text-to-image models typically utilize an encoder-decoder architecture, where encoders learn the contextual representations of input text, and decoders use the learned information to generate corresponding images.
Particularly, text encoders are usually pre-trained language models (\feng{\eg} GPT~\cite{brown2020language} and BERT~\cite{DevlinCLT19}), and to-image decoders generally use GAN-based and Diffusion-based models.

GAN-based models \cite{ZhuP0019, zhang2017stackgan} contain a generator and a discriminator, where the generator accepts text encodings and generates output while the discriminator tries to differentiate the output from real image examples.
Diffusion-based models~\cite{rombach2022high, ramesh2022hierarchical, nichol2021glide} learn to remove noise from random images and generate final images that match the text information. 
For instance, Stable Diffusion~\cite{rombach2022high} involves a latent diffusion process where the model learns to remove noise from the random noised images in the embedding space with the guidance of text input. The denoising process leads to high-quality images with state-of-the-art performance.
% With its high performance and availability to the public, 

However, text-to-image generation quality greatly depends on natural language prompts and human subjective judgment.
It requires humans in the loop to refine the generation~\cite{wong_dpviscreator_2023}.
Although there are some open-sourced demos, such as Stable Diffusion\footnote{\url{https://huggingface.co/spaces/stabilityai/stable-diffusion}}, Midjourney\footnote{\url{https://www.midjourney.com/}}, and DALL$\cdot$E 2\footnote{\url{https://openai.com/product/dall-e-2}}, for the public to create their own artwork with natural language input, users need to try different phrasings to derive the desired output, which can be time-consuming. 
In this paper, we propose an interactive visual analytics system that can summarize and recommend prompt keywords to help formulate and refine users' prompts based on an external large text-to-image prompt dataset, DiffusionDB\cite{wang2022diffusiondb}.

%% file: chapters/3-design-process.tex
\section{Overview}
\subsection{Background}
\textbf{Stable Diffusion.}
Based on the observations of particle diffusion in physical systems and modeling of the inverse process~\cite{sohl2015deep}, denoising diffusion probabilistic models achieved significant improvement in generating high-resolution images~\cite{ho2020denoising, song2020denoising}. 
For diffusion models, a forward process is defined by a series of steps for adding noise to the image, and the corresponding backward process (\ie the denoising process) is modeled by deep neural networks. 
The denoising process can be guided by extra conditions, such as text inputs (\ie text prompts).
The importance of the text prompt guidance is controlled by the hyper-parameter \textit{guidance scale}.
A larger guidance scale brings better alignments between the generated images and the prompts with the sacrifice of image diversity.
Stable Diffusion \cite{rombach2022high}, one of the state-of-the-art diffusion models, achieves high performance while consuming fewer computational resources.
By compressing the images from pixel space into latent space, Stable Diffusion preserves the semantic information while removing the image details, resulting in a simplified representation space and a faster generation process.
Our study utilizes Stable Diffusion for image generation, and it can be replaced by other text-to-image generative models for specific applications.

\textbf{DiffusionDB.}
The popularity of Stable Diffusion has led to a surge in individuals sharing their image creations and input prompts on public social platforms.
This trend has sparked new studies aimed at collecting and analyzing publicly shared results for future research opportunities.
DiffusionDB \cite{wang2022diffusiondb} is the first large-scale dataset that comprises 14 million input-output data pairs (\ie text prompts and hyper-parameters input by users and their corresponding model-generated images).
DiffusionDB anonymizes image creators to protect user privacy and excludes harmful or NSFW \feng{(not safe for work)} images.
Since some users may use the same text prompt for several attempts with different hyper-parameters (\eg random seeds and guidance scales), the 14 million data items contain 1.8 million unique text prompts.
To facilitate prompt feature analysis, DiffusionDB also offers a subset version called DiffusionDB-2M, which includes 1.5 million unique text prompts and their corresponding 2 million generated images.

\begin{figure*}[ht]
 \centering % avoid the use of \begin{center}...\end{center} and use \centering instead (more compact)
 \includegraphics[width=\linewidth]{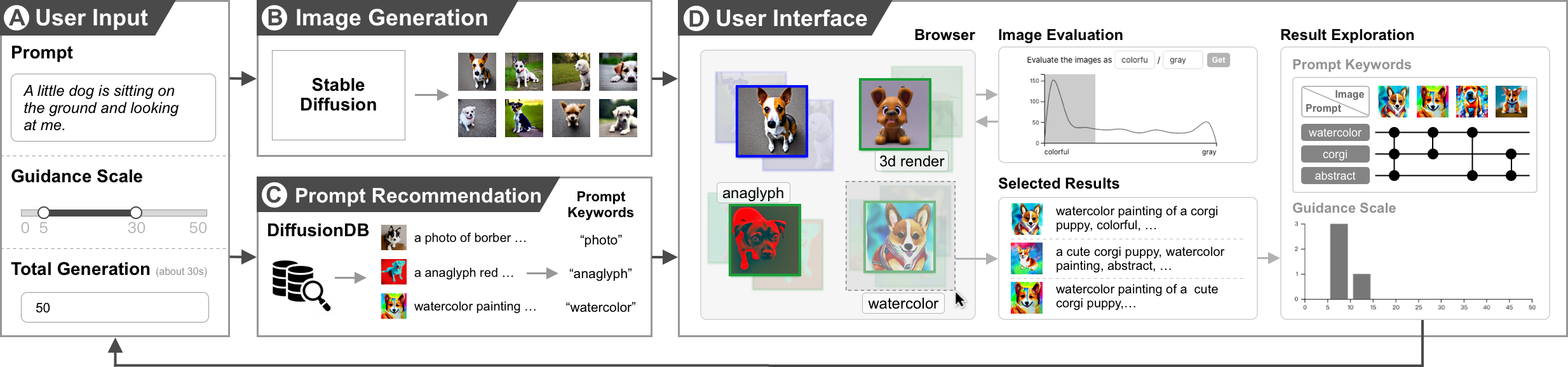}
 \caption{The {\name} framework consists of four major components. It enables users to (A) specify model input for text-to-image creation. {\name} (B) generates a set of images using Stable Diffusion and (C) identifies related prompt keywords for recommendations. (D) Both the image results and prompt keywords are visualized in the user interface to support interactive exploration for prompt engineering.}
 \label{fig:framework}
\end{figure*}

\subsection{Design Requirements}
The target users of our study are ordinary users who are interested in image creation but lack the expertise to use professional tools.
Often, these users struggle to produce high-quality images that meet their expectations.
A potential solution to this problem is the utilization of a text-to-image generative model.
The primary objective of our study is to design a system that facilitates collaboration between the users and the generative model.
% Participants
We recruited 9 ordinary users (P1-P9) who are interested in text-to-image creation from local universities.
P1-P4 are familiar with interactive tools or online demos for image creation, and others are aware of such tools but are not very familiar with them.

% Procedure
To identify the design requirements of the system, we interviewed the participants during the early stage of the study.
In the interview, participants were asked to create images using publicly available systems, including the online demo of Stable Diffusion and Lexica \footnote{\url{https://lexica.art}}, an online website system that supports similar image-prompt search.
We encouraged participants to engage in open-ended exploration without the constraints of content or style.
Then, we conducted interviews with participants to collect their feedback and comments regarding the usage of text-to-image models.
Specifically, we focused on how the participants get the expected results and how to facilitate this process.
In the following four months, we conducted regular meetings with them to update the design requirements and gain feedback to guide the design and development of our system prototype.
Finally, the design requirements of our system were summarized as follows.

\textbf{R1. Generate a collection of image results for the user prompt.}
In some online demos of text-to-image tools, the users can only get a few results per prompt submission and have to manually try different model inputs (\ie text prompt and model hyper-parameter) each time to get more image results.
It may be time-consuming for the users to get the desired results.
The system should help users efficiently get a collection of image results for exploration.

\textit{R1.1. Generate multiple image results with varying hyper-parameter values.}
For the same user prompt, the Stable Diffusion model can generate different image results using different hyper-parameters, including the guidance scale and random seed.
The users often encounter confusion when receiving undesirable image results, as it is challenging to assess whether the prompt itself is inadequate or requires better model hyper-parameters.
The system should allow users to specify multiple hyper-parameter values to generate multiple images at once, which helps users efficiently evaluate the quality of prompts.

\textit{R1.2. Provide similar image work to inspire prompt refinement.}
The target users of our system do not have to be familiar with the model's architecture and prompting strategies.
By exploring and comparing the image works and their prompts, users can gain insight into what kinds of results the model is capable of generating and how to phrase or refine their prompt to obtain such results.
The system should provide previous image works that are similar to user prompts and help users gain inspiration for prompt refinement.

\textbf{R2. Provide a visual summary for image collection.}
The purpose of image exploration is to find image results of interest to inspire prompt refinement.
However, exploring a large image collection is a time-consuming process.
The system should provide a visual summary of the image collection so that users can easily overview the image characteristics and navigate to the image subset for detailed exploration.

\textbf{R3. Support efficient image evaluation from different aspects.}
The images in the collection may have diverse subjects or styles.
Users desire to specify evaluation criteria for images from different aspects (\eg image subject, color style, and visual perception) and automatically evaluate the image results so that users can gain an overview of the image distribution in terms of these aspects and focus on the image subset of interests for efficient exploration.

\textbf{R4. Support iterative refinement of prompt and model hyper-parameters.}
Given the flexibility of natural language and the subjective nature of image creation, users usually need to iteratively refine the model input, including prompt and model hyper-parameters (\ie guidance scales and random seeds), to get the desired results.
However, it is challenging to identify effective prompt keywords.
The system should recommend prompt keywords for the related images and support the joint exploration of prompt keywords and their corresponding images for validation.
Moreover, the system should help users explore the model hyper-parameters of the images of interest.

\subsection{System Overview}
The workflow of our system is summarized in \autoref{fig:framework}.
Our system supports user input of prompts and model hyper-parameters (\textbf{R1}), including the range of guidance scale and the number of generations (for different random seeds).
Then the system generates a collection of images using prompts and hyper-parameters (\textbf{R1.1}).
To help users improve the prompts, the system introduces a prompt recommendation model that retrieves similar creation results from DiffusionDB (\textbf{R1.2}) and identifies the related prompt keywords from the corresponding prompts.
Both the model-generated and retrieved images and recommended prompt keywords are co-embedded into 2D space according to their semantics and presented in multi-level visualization to facilitate exploration (\textbf{R2}).
Based on that, the system enables users to specify the aesthetic evaluation criteria (\eg beauty) to efficiently evaluate and select the images (\textbf{R3}).
The users can select image subsets of interest for further exploration of their prompt keywords and guidance scales, which can be used to refine the user input (\textbf{R4}).

%% file: chapters/4-method.tex
\section{Prompt Recommendation}
\begin{figure*}[ht]
 \centering % avoid the use of \begin{center}...\end{center} and use \centering instead (more compact)
 \includegraphics[width=\linewidth]{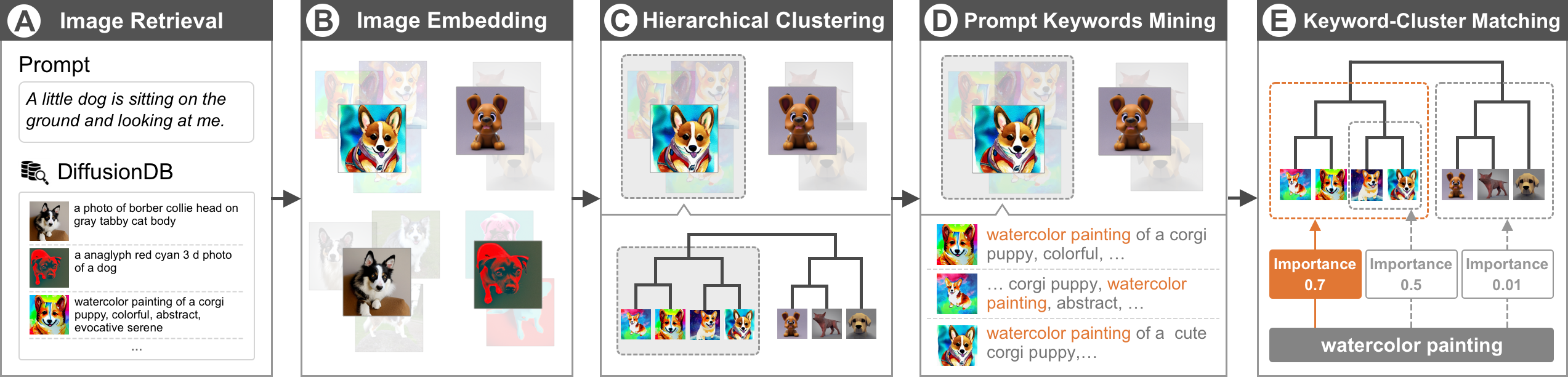}
 \caption{The pipeline of the prompt recommendation model involves five steps: (A) retrieving similar images from the DiffusionDB dataset; (B) embedding them according to their semantics; (C) arranging them into hierarchical clusters; (D) mining special and related keywords from the prompts in the clusters; and (E) matching each keyword with its most related cluster.}
 \label{fig:approach}
\end{figure*}

The prompt recommendation model mines special and relevant prompt keywords from similar image creations.
As shown in \autoref{fig:approach}, the model pipeline consists of five steps: (A) retrieving image results similar to user input prompts from the DiffusionDB dataset; (B) embedding images according to their semantic features; (C) conducting hierarchical clustering of images; (D) identifying important and special prompt keywords from image clusters; and (E) matching each prompt keyword to its most related image cluster.

\subsection{Image Retrieval}
\feng{To retrieve similar images that match the user prompts, both images and their original prompts in the DiffusionDB dataset~\cite{wang2022diffusiondb} can be used as a search space.
However, due to the nature of text-to-image generative models~\cite{liu2022design}, many cases in the DiffusionDB contain significantly different images generated by the same or very similar prompts~\cite{wang2022diffusiondb}.
Using the image features as the search space can better distinguish their differences and return more similar results.}
To align user prompts with the image space, we utilize the CLIP model \cite{radford2021learning}, a state-of-the-art model in contrastive representation learning.
Pre-trained on a vast dataset with 400 million text-image pairs, the CLIP model has aligned the feature vectors of the text and image in each pair.
We use cosine distance to measure the feature similarity of user prompts and images.

\subsection{Image Embedding}
For the retrieved results, we utilize both the images and their prompts for embedding to better capture the semantic features of the images, such as the subjects and style.
Similar to the encoding schema for image retrieval, we employ the CLIP model to encode both the images and prompts into 512-dimensional vectors.
Then, the text and image features are concatenated together as \feng{512+512=1024} dimensional vectors, which serve as the final representation of the images.
Through this process, we can uncover the semantic similarity of images based on their distance in the 1024-dimensional space.
\feng{Compared to individual text or image features, the concatenated features can better aggregate similar images with similar prompts.}
To enable user exploration of images according to their semantic similarity, we employ the t-SNE algorithm \cite{van2008visualizing} for feature dimension reduction.
We set the cosine distance as the \textit{metric} parameter of the t-SNE algorithm, aligning it with the training objective of the CLIP model.

\subsection{Hierarchical Clustering}
Based on the representation features of images that reveal their semantic similarity, we organize the images with a clustering algorithm to aggregate common characteristics (\eg image colors and styles).
Since there are no discrete labels available for the retrieved images, it is hard to determine the number of clusters for images, which is an important parameter that needs to be pre-specified for some clustering algorithms, like k-means \cite{macqueen1967classification}.
Therefore, we use hierarchical clustering, which models the cluster structure as a tree.
As shown in \autoref{fig:approach}C, each image (leaf nodes in the tree) is initialized as its own cluster and progressively merged with the neighboring nodes into a larger cluster (non-leaf nodes in the tree).
The clustering process is bottom-up and ends when all the image nodes are merged into one tree.

Each non-leaf node in the tree denotes an image cluster in the embedding space, but not all clusters are suitable for prompt keyword mining, as we aim to identify the prompt keywords that are special and strongly associated with the image clusters.
Overly large clusters contain many generic keywords ({\eg} stop words) and obscure specific ones ({\eg} magical words \cite{wang2022diffusiondb}).
We constrain the volume of clusters based on the number and position range of child nodes.
We limit the range of the number of child nodes from 3 to 20 according to the total number of retrieved images.
For the positional constraints \feng{in the 2D space}, we discard clusters merged from relatively distant sub-clusters, which are not appropriate for prompt keyword mining.

\subsection{Prompt Keyword Mining}
In line with prior studies \cite{liu2022design, oppenlaender2022taxonomy} that emphasize the importance of prompt keywords (\ie prompt modifiers and phrases) over connecting words (\ie sentence structure), we focus on recommending prompt keywords to help users refine the original prompt input.
For each cluster, we aim to identify special keywords from the prompts of the image clusters.
By ``special,'' we mean that these keywords are significantly more crucial for the current cluster than for others.
To measure the importance of each keyword for the current cluster, we compute the TF-IDF values of keywords at the cluster level:
\begin{equation} \mathrm{tfidf_{i,x}} = \mathrm{tf_{i,x}} \times \mathrm {idf_{i}} \end{equation}
where the $\mathrm{tf_{i,c}}$ is the term frequency of the given keyword $t_{i}$ in the current cluster $c_{x}$.
It measures the importance of the keyword for the current cluster and is calculated as follows:
\begin{equation} \mathrm{tf_{{i,x}}} = \frac{n_{i,x}}{\sum_{k}n_{k,x}} \end{equation}
the $n_{i,x}$ is the number of keyword $t_{i}$ in the current cluster $c_{x}$ which consists of multiple prompt documents $d_{{j}}$:
\begin{equation} n_{i,x} = \sum_{d_{j} \in c_{x}} n_{{i,j}} \end{equation}
The $\mathrm{idf_{i}}$ is the inverse document frequency of this keyword which measures the inverse importance of the keyword for the whole prompt set of the retrieved images.
\begin{equation} {\mathrm{idf_{i}} = \lg{\frac {|D|}{|\{j:t_{i}\in d_{j}\}|}}} \end{equation}
where the $|D|$ is the total number of prompts for the retrieved image collection and ${|\{j:t_{i}\in d_{j}\}|}$ is the number of prompts containing the given keyword $t_i$.
Consequently, generic keywords tend to receive higher $\mathrm{tf}$ values but lower $\mathrm{idf}$ values, which renders them less likely to be identified as the most special keywords for the cluster. 

Moreover, since the special prompt keywords can comprise multiple words \cite{wang2022diffusiondb}, such as ``\textit{unreal engine}'' and ``\textit{trending on Artstation},'' we incorporate \textit{n-grams} to detect special multi-word phrases in the prompts.
After computing the importance value of the keywords in the cluster, we eliminate the stop words (\eg ``\textit{the}'' and ``\textit{and}'') using the NLTK toolkit \cite{loper2002nltk}.
Please note that the stop words in the \textit{n-gram} (\eg ``\textit{on}'' in ``\textit{trending on Artstation}'') are excepted since they are connecting words for the other words.
Finally, we select the top prompt keywords for each cluster according to their importance values.

\subsection{Prompt-Cluster Matching}
Following prompt keyword mining, the keywords may occur in multiple clusters, each with varying levels of importance.
\feng{Matching the prompt keywords to their most related cluster node can better illustrate the effect of the keywords~\cite{wong_cohortva_2023} (\ie what can be generated by the prompt keywords).}
This mapping is especially beneficial when visualizing the images and text jointly to aid user exploration and comprehension \cite{xie2018semantic}.
To accomplish this, we normalize the importance values of keywords within the same cluster and select the cluster with the highest TF-IDF value for the given keyword $t_i$ as its best cluster $cb_{t_i}$.
\begin{equation} cb_{t_i} = \underset{c_x\in c_{t_i}}{\arg\max}(\mathrm{tfidf_{i,x}}) \end{equation}
Here, $c_{t_i}$ denotes the set of clusters $c_x$ that contain the keyword $t_i$.
To mitigate redundancy, we combine \textit{n-gram} keywords belonging to the same cluster.
For example, the keywords ``\textit{unreal},'' ``\textit{engine},'' and ``\textit{unreal engine}'' are associated with the same cluster with similar importance, indicating that the two individual words are typically used together.
Thus, we retain the ``\textit{unreal engine}'' and eliminate the two individual words ``\textit{unreal}'' and ``\textit{engine}.''

%% file: chapters/5-user-interface.tex
\begin{figure*}[ht]
 \centering % avoid the use of \begin{center}...\end{center} and use \centering instead (more compact)
 \includegraphics[width=\linewidth]{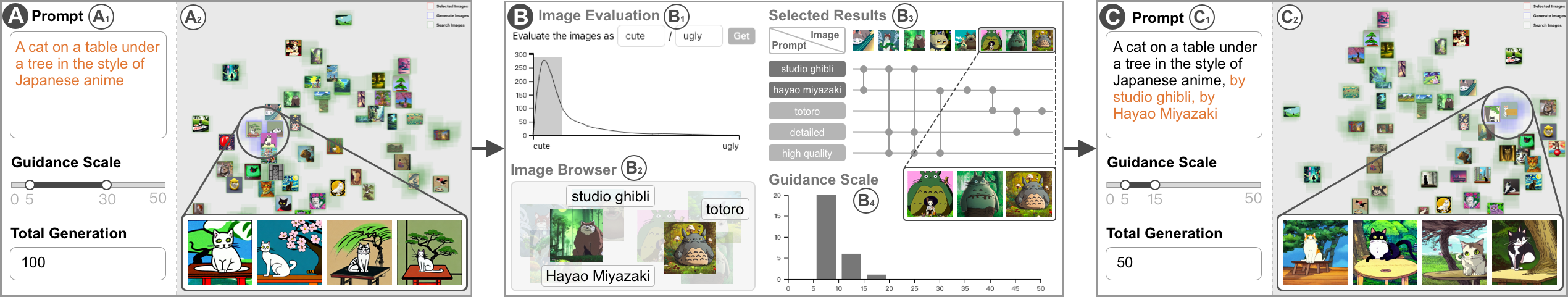}
 \caption{The first usage scenario showcases how our system supports the user in refining the prompt for the desired image style. (A) The user tries the first prompt but is not satisfied with the model-generated images. (B) She then explores similar images and receives recommended prompt keywords that clarify the desired image style. (C) Finally, she successfully obtains the desired results on the second attempt.}
 \label{fig:case_one}
\end{figure*}

\section{System Design}
We have developed a visual analysis system that leverages the Stable Diffusion model and our prompt recommendation model to assist users in interactive prompt engineering and text-to-image creation.
In this section, we first describe the user interface of {\name}, followed by a detailed explanation of critical system components, including the multi-level image-prompt visualization and the image evaluation.

\subsection{User Interface}
The user interface (Figure 1) encompasses four views.
The \textit{Model Input View} (Figure 1A) allows users to input the text prompt and customize the model hyper-parameters.
The \textit{Image Browser View} (Figure 1B) visualizes the image collection, including model-generated and retrieved images as well as the recommended prompt keywords.
The \textit{Image Evaluation View} (Figure 1C) empowers users to establish aesthetic criteria for assessing and filtering out irrelevant images.
Users can navigate through the \textit{Image Browser View} and select specific images for further examination in \textit{Local Exploration View} (Figure 1D).

\textbf{Model Input View} serves as the starting point of the image creation process.
Users can input a prompt to describe the desired subjects and styles and configure the model hyper-parameters, including guidance scale and total generation.
Once the range of the guidance scale is specified, the system randomly samples values within the range.

\textbf{Image Browser View} is the primary window for users to explore the image collection.
Both the generated and retrieved images and the prompt keywords are co-embedded and visualized within the view.
The images are positioned based on their semantic embedding (Section 4.2), and the keywords are positioned near their most related image clusters (Section 4.5).
The images and keywords are presented in multi-level visualization to reduce the visual clutter~\cite{bertucci2022dendromap, wong_taxthemis_2021}.
At the overview level, representative images for the clusters are visualized, and the others are replaced by translucent rectangles.
As the users navigate to the detailed level, all images within the clusters and their corresponding prompt keywords are gradually revealed.
\feng{Users can hover over an image to view its detailed information or click a keyword to highlight specific images containing it.}
The title bar of \textit{Image Browser View} includes checkboxes for controlling the visibility of the image types (\ie generated or retrieved) and keywords.

\textbf{Image Evaluation View} allows users to specify criteria for filtering images from multiple aspects ({\eg} aesthetic).
Users can input two keywords (\eg \textit{simple} and \textit{complex}) that represent two poles of an evaluation criterion (\eg image complexity).
\feng{The second keyword is optional and will be set as the opposite of the first keyword using ``not'' if not specified.}
The system then rates all images based on the two keywords (Section 5.3) and visualizes the rating distribution to support brush interaction for image filtering.
At the top of the view, we recommend some common pairs of keywords for image evaluation \cite{wang2022exploring}.

\textbf{Local Exploration View} comprises three panels.
The \textit{Result Details Panel} presents detailed information about the selected results, enabling users to view the high-resolution images and explore their prompts and hyper-parameter values.
\feng{The \textit{Prompt Keywords Panel} not only presents the prompt keywords (sorted by the keyword importance introduced in Section 4.4) for the selected images but also visualizes their interrelationship with the images using BioFabric~\cite{longabaugh2012combing}, a tabular layout for graph visualization.
Each point in the table indicates that a keyword is used by an image.
The users can explore how frequently the keywords are used by the selected images, or compare the images with different keywords to better understand the effect of the keywords.
The users can also select the keywords to highlight them in their context in \textit{Result Details Panel} for easy location.}
The \textit{Guidance Scale Panel} visualizes the distribution of the guidance scales of the selected images in a histogram.
Both the recommended prompt keywords and the range of the guidance scale can assist users in refining their original input to obtain better results from the Stable Diffusion model.

\subsection{Multi-Level Image-Prompt Visualization}
All generated and retrieved images are projected into a 2-D space with the t-SNE dimensional redundancy algorithm.
To minimize visual clutter at the overview level and \feng{support semantic zoom~\cite{PerlinF93, BuringGR06, LYiWLG22, abs-1906-05996}}, we employ the hierarchical structure of image clusters to construct multi-level visualization of images.
For the image clusters, we choose the images closest to the cluster center as the representative images and present them at the overview level.
For the prompt keywords mapped to the image clusters, we position them near the images whose prompts contain the keywords.
The positions of the keywords $p_{t_i}$ are calculated by the weighted average of their image positions:
\begin{equation} p_{t_i} = \frac{\sum n_{i,j} \times p_j }{\sum n_{i,j}} \end{equation}
Here, $p_j$ donates the position of the images, and $n_{i,j}$ donates the number of times the keyword appears in the prompt of the images.
If multiple keywords are positioned in the same position, we add a small random shift to the position of the keyword to avoid overlap.
The prompt keywords are also visualized in multi-level, corresponding to the level of their respective clusters.

\subsection{Image Evaluation}
Image evaluation allows users to specify criteria for images \feng{from both objective aspects \cite{radford2021learning} ({\eg} plane, cat, and dog) and subjective aspects ({\eg} quality perception \cite{CheonYKL21, YeKKD12, ZhangZMZ11} and abstract perception \cite{wang2022exploring, MaYY017, AchlioptasOHEG21, PandaZLLLR18}).}
Inspired by prior work~\cite{wang2022exploring}, we use CLIP model~\cite{radford2021learning} to capture the relationship between text and visual perception.
We use pairs of opposing texts related to human perception (\eg \textit{real} and \textit{abstract}) to fill in the pre-defined template for image evaluation (\ie [text] image).
We then calculate the feature cosine similarity $s_i$ of each image with the two texts and compute the image rating (represented by $\bar{s}$) between the pair of keywords on a scale of $[0,1]$ using Softmax:
\begin{equation} \bar{s}=\frac{e^{s_1}}{e^{s_1}+e^{s_2}} \end{equation}
Utilizing two opposing keywords transforms the image evaluation task to a binary classification, which can effectively reduce the ambiguity that arises from using a single keyword \cite{wang2022exploring}.
The closer $\bar{s}$ is to 0 or 1, the closer the image is to the keyword $t_1$ or $t_2$.
\feng{Sometimes it is not easy for users to specify the second opposing keyword. We make it optional and generate the opposite of the first keyword using the negative word ``not.''
The image evaluation strategy can be further extended to support natural language sentences to better distinguish the nuances in the images, and we leave that for future work.}

%% file: chapters/6-case-study.tex
\begin{figure*}[t]
 \centering % avoid the use of \begin{center}...\end{center} and use \centering instead (more compact)
 \includegraphics[width=\linewidth]{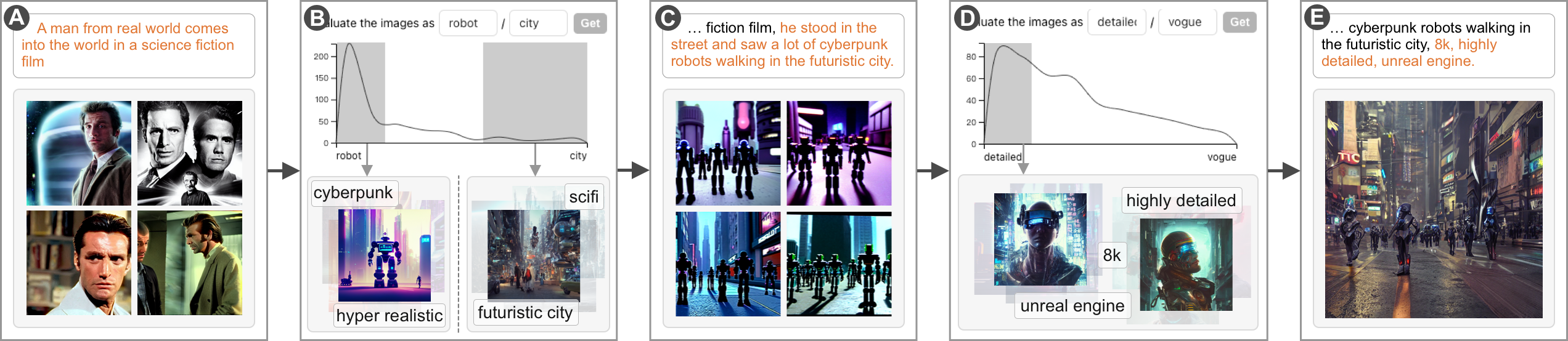}
 \caption{The second usage scenario presents how our system facilitates open-ended creation. (A) The user inputs the first prompt and gets unexpected image results. (B) So he overviews the image collection and explores the prompt keywords for city and robot images separately. (C) Then he revises the original prompt and gets a new collection of images. (D) To improve the details of the images, the user specifies criteria to select highly detailed images. (E) With the recommended keywords, the user finally gets a satisfied image result.}
 \label{fig:case_two}
\end{figure*}

\section{Usage Scenarios}
In this section, we present two usage scenarios that showcase the utility of our system.
The first usage scenario demonstrates how our system helps the user improve her text prompt and adjust the model hyper-parameters to achieve the desired image result with the target image style.
The second usage scenario exemplifies how our system inspires the user with a broad creative goal and guides him in gradually clarifying the subjects of the images and improving the quality of the generated results.

\subsection{Scenario 1: Prompting for the Desired Image Style}
In this scenario, the user desires to create an anime-style work featuring her cat.
She provides the initial prompt ``\textit{a cat on a table under a tree in the style of Japanese anime.}'' in \textit{Model Input View} (\autoref{fig:case_one}$A_1$).
Being unfamiliar with the model hyper-parameters, she sets the guidance scale range from 5 to 30 and requires the system to generate 100 images at a time.
Following a brief waiting period, the system produced a collection of generated images in the \textit{Image Browser View} (\autoref{fig:case_one}$A_2$).
After browsing through the results, the user finds that the generated results do not align with her expectations.

To refine her prompts toward the desired style, the user explores similar images for inspiration.
Given the considerable number of search results, the user specifies artistic criteria based on the keyword pair ``\textit{cute}'' and ``\textit{ugly}'' to select images (\autoref{fig:case_one}$B_1$).
In response, the system visualizes the distribution of all images' ratings on this criteria.
The user brushes to select the most ``cute'' images.
From the remaining images in the \textit{Image Browser View}, she identifies some images characterized by a Totoro style (\autoref{fig:case_one}$B_2$).
The presence of prompt keywords surrounding the images confirms her finding.
As the user prefers this style, she navigates the detailed level of the view and selects these images using a brush interaction for deeper exploration.

The details of the selected images are listed in the \textit{Local Exploration View}.
After a brief review of the results, the user proceeds to examine the suggested prompt keywords in the \textit{Prompt Keywords Panel} (\autoref{fig:case_one}$B_3$).
\feng{She finds that the top-recommended prompt keywords ({\ie} ``\textit{Hayao Miyazaki},'' ``\textit{Studio Ghibli},'' and ``\textit{Totoro}'') are strongly associated with the style and frequently used by the selected images.
However, the ``\textit{Totoro}'' keyword has a more substantial impact on the subject of the images, which is not what the user expects.
Therefore, the user discards ``\textit{Totoro}'' and chooses the other two keywords to refine her original prompt (\autoref{fig:case_one}$C_1$).}
Additionally, the user specifies a narrower range of guidance scales (5-10), referencing the selected images (\autoref{fig:case_one}$B_4$).
As a result, the system generates a collection of images in the style of the Totoro movie (\autoref{fig:case_one}$C_2$).
Finally, the user selects a favorite image as the outcome of this creation process.

\subsection{Scenario 2: Prompting for Open-Ended Creation}
In this scenario, the user begins with the broad creative goal of illustrating a picture of a future world.
He inputs the prompt ``\textit{a man from the real world comes into the world in a science fiction film}'' to leverage the imaginative capabilities of the generative model (\autoref{fig:case_two}A).
However, the model-generated results primarily consist of freeze-frames of male characters from old science fiction films, which are not the intended subjects.
Therefore, the user explores similar retrieved images to refine the image subjects.

Upon reviewing the overview of the \textit{Image Browser View}, the user recognizes that the retrieved images primarily feature individuals (robots or cyborgs) or cities, inspiring the user to create images using these subjects.
To explore these subjects and their corresponding keywords further, the user specifies a pair of keywords, ``\textit{robot}'' and ``\textit{city},'' for image selection (\autoref{fig:case_two}B).
Using the rating distribution chart, the user then successively explores the images ``close'' to robots and cities through brush interaction.
With the help of recommended prompt keywords, the user improves his original prompt by adding a second sentence that better clarifies the image subjects ``\textit{he stood in the street and saw a lot of cyberpunk robots walking in the futuristic city.}''

For the image results of the second generation (\autoref{fig:case_two}C), the user notices that the city scene has a futuristic feel, but the robots lack sufficient detail and appear vague.
Therefore, the user continues to explore the retrieved images that are ``close'' to robots and specify additional criteria for image quality using the keywords ``\textit{detailed}'' and ``\textit{vogue}'' (\autoref{fig:case_two}D).
Upon examining some detailed robot images, the user identifies a set of prompt keywords and phrases, such as ``\textit{8k},'' ``\textit{highly detailed},'' and ``\textit{unreal engine}.''
He adds these keywords and phrases to his prompt (\autoref{fig:case_two}E).
This time, the results returned contain more images with high details and textures, from which the user selects an image as the final creation outcome.

%% file: chapters/7-evaluation.tex
\section{User Study}
We conducted a user study to evaluate the effectiveness and usability of our system in facilitating interactive prompt engineering and image creation.
Specifically, we aim to evaluate (1) the helpfulness of the prompting recommendation model, (2) the effectiveness and usability of the overall system, and (3) the creativity support compared with two baseline systems that mimic real-world text-to-image creation.

\subsection{Participants}
% Participants
\feng{We recruited 12 participants (P1-P12, four females and eight males, aged 24-32) from a local university through an internal school forum.
The participants are mainly undergraduate and master's students from various disciplines, including industrial design, digital media, computer science, and literature.}
They had more or less experience with text-to-image generation tools but lacked sufficient knowledge about the generative models and how best to use them.

\subsection{Baseline Systems}
We designed two baseline systems for comparative study alongside our system.
All three systems utilize the Stable Diffusion model as the backbone model for text-to-image creation.

\textbf{Baseline A} provides the same similar image retrieval as our system without the prompt keyword recommendation feature.
This baseline mimics a typical creation scenario where users search related artwork created by previous authors on public platforms (\eg Lexica) to gain inspiration for their own text-to-image creation.

\textbf{Baseline B} presents only the model-generated images in the \textit{Image Browser View} without the image retrieval and prompt keyword recommendation.
However, it introduces Promptist \cite{hao2022optimizing}, an automatic prompting method that automatically helps users refine the prompts to improve the aesthetic quality of generated images.

\subsection{Procedure and Tasks}
\textbf{Introduction (15 min).} 
We first provided a brief introduction of the research background, including the research motivation and the study protocol.
Next, we gathered demographic information from the participants and asked for their consent to record their operations and results for further analysis.
We then introduced the feature of the system views and demonstrated their usage with clear examples~\cite{wong_example_2023}.

\textbf{Target replication training (15 min).}
To help users become familiar with the systems, we set a training task that requires the participants to replicate some given images (\eg the images in \autoref{fig:case_two}E) using our system and Promptist.
Following the Jeopardy evaluation methodology \cite{gao2015datatone}, we did not provide textual descriptions about the target images and the metadata (\eg guidance scales and random seeds).

\textbf{Open-ended creation (60 min).}
In this stage, the participants were required to conduct open-ended creation tasks under a specific theme ({\eg} city) without detailed constraints on the target of the creation, which aimed to evaluate the overall features and usability of the systems \cite{chen2022crossdata, srinivasan2020ask, feng_xnli_2023}.
The participants were required to utilize three systems (up to 20 minutes per system) for image creation.
\feng{For each system, the participants needed to (1) choose a theme for image creation, (2) write a prompt for the initial attempt, and (3) iteratively improve the prompts using the system.}
The order of the systems was counterbalanced.

\begin{figure}[t]
 \centering % avoid the use of \begin{center}...\end{center} and use \centering instead (more compact)
 \includegraphics[width=\linewidth]{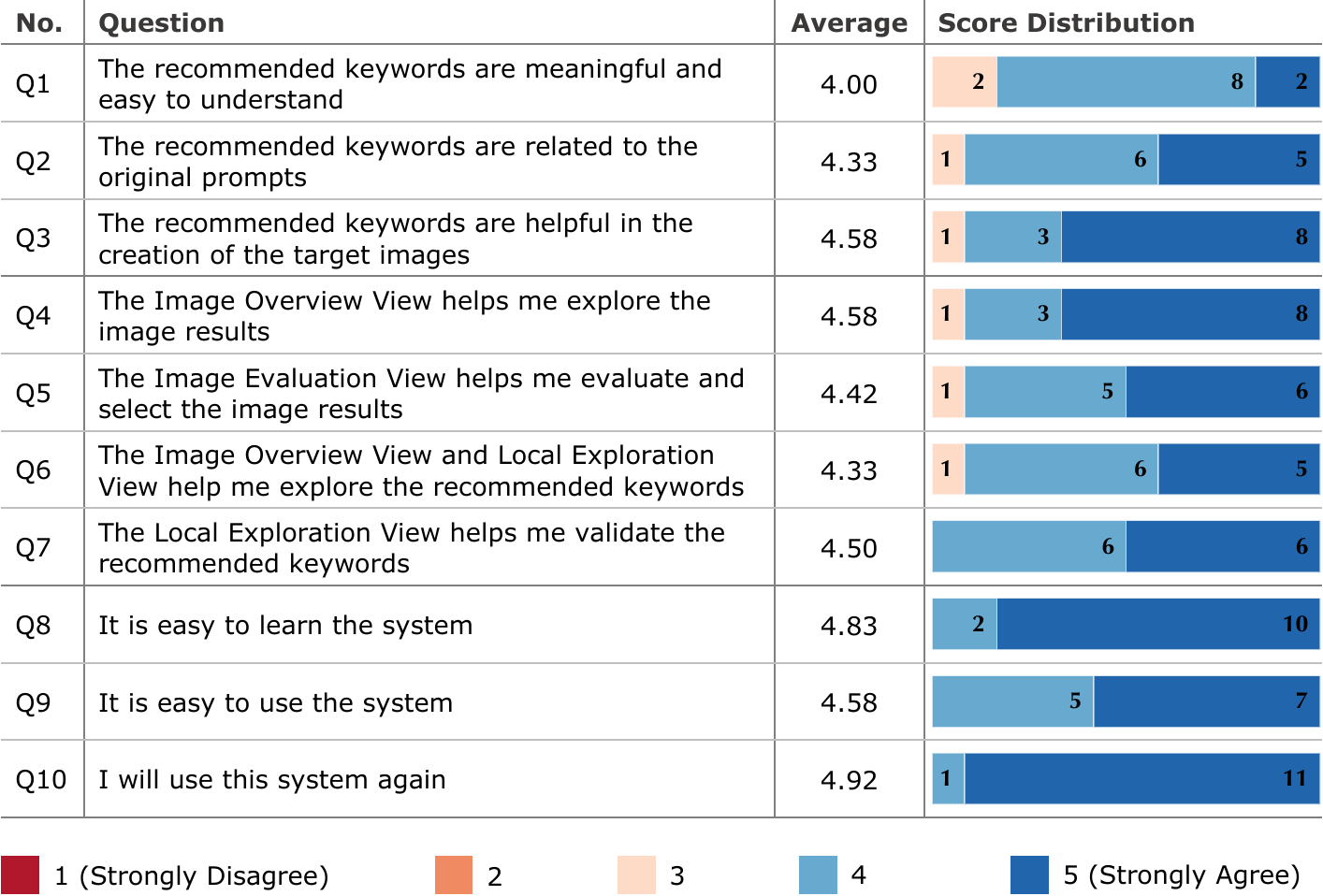}
 \caption{The results of the questionnaire regarding the effectiveness and usability of the visual system and prompt recommendation model.}
 \label{fig:effective}
\end{figure}

\textbf{Semi-structured interview (30 min).}
We asked the participants to complete a five-point Likert-scale questionnaire to evaluate the effectiveness and usability (\autoref{fig:effective}) of our system and the creativity support of all three systems (\autoref{fig:comparison}).
Finally, we conducted an interview with the participants to collect their feedback for further analysis.

\subsection{Results Analysis}
All the participants completed the training and open-ended creation tasks and experienced the system's functions.
Based on the user ratings in the questionnaire and feedback received during the interview, we discuss the effectiveness and usability of the system and prompting model. We also report the difference between our system and the other two baselines.
Finally, we report insights into user patterns when using our system for prompt engineering as well as areas for improvement.

\subsubsection{Effectiveness of the Prompting Model}
Most of the participants thought the recommended keywords were \textbf{meaningful} and \textbf{easy to understand} (Q1).
\feng{P3 commented, ``\textit{I can easily understand which parts of the prompts I can improve with the keywords.}'' Sometimes the participants were not familiar with some artists' names, such as ``\textit{Leonid Afremov}'' (P8).}
Most of the participants agreed that the recommended prompt keywords were \textbf{related} to their original prompts (Q2).
\feng{P6 found that the prompt keywords might be directly related ({\eg} the style or composition of the images) or indirectly related ({\eg} commonly paired objects in similar images) to the prompts.
P7 pointed out that some recommended keywords were for general purposes and were not related to specific subjects, such as ``\textit{highly detailed}.''}
The recommended keywords were considered to be \textbf{helpful} in the prompt improvement (Q3) regarding stylization and image quality.
\feng{For instance, P10 appreciated that the recommended keywords ``\textit{summary afternoon}'' helped him better control the hue of the sky in the images. P4 felt impressed that the recommended keyword ``beautiful'' significantly improved the aesthetic qualities of the images.}

\subsubsection{Effectiveness of the Visual System}
The \textit{Image Browser View} was appreciated by the participants for \textbf{image exploration} (Q4).
\feng{The semantic-based visualization ``\textit{effectively grouped similar images together}'' (P3) and facilitated users to ``\textit{batch select images for detailed comparison}'' (P7).
The multi-level visualization further improved the exploration experience as it ``\textit{provided natural switching between the levels}'' (P4) and ``\textit{reduced cognitive load}'' (P9).}
The \textit{Image Evaluation View} helped most participants \textbf{evaluate and select images} (Q5) \feng{as it supported free-from criteria, which helped the users ``\textit{narrow down the exploration space and accelerate the exploration process}'' (P1).
It was interesting that it even worked for some special aspects, like the strength (strong or weak) of the animals in the images.}
The \textit{Image Browser View} and \textit{Local Exploration View} were found to be useful for \textbf{exploring the recommended keywords} (Q6).
\feng{The participants confirmed that visualizing the keywords near the images helped them understand both.
In addition, highlighting images around keywords ``\textit{showcased the impact scope of the keywords}'' (P1) and ``\textit{provided hints on image selection}'' (P8).
For the \textit{Prompt Keywords Panel}, the participants appreciated visualizing the correlation between keywords and images. P1 commented that it helped discover common keyword combinations and what could be generated using them.}
Most participants confirmed that the \textit{Local Exploration View} helped them \textbf{validate the recommended keywords} (Q7).
\feng{P8 mentioned that comparing the images containing different keywords was quite useful to exclude keywords that were frequently used but not related to the expected aspects.
The highlighting of the keywords in their original prompts also helped users ``\textit{understand the context of keywords}'' (P9).}

\subsubsection{Usability}
All the participants agreed that our system was \textbf{easy to learn} (Q8) and \textbf{easy to use} (Q9).
\feng{The participants thought that the workflow of our system was intuitive and the interface was user-friendly.
They also appreciated the system's reminders of estimated times for image generation, which helped them make better trade-offs between the number of images generated and waiting time.
P6 suggested adding a mini-map for the \textit{Image Browser View} to better support navigation.}
Lastly, all the participants expressed their willingness to \textbf{use our system again} for creation tasks in the future (Q10).
\feng{P12 stated, ``\textit{Exploring the images in the system was immersive, and I enjoyed the process.}''}

\begin{figure}[ht]
 \centering % avoid the use of \begin{center}...\end{center} and use \centering instead (more compact)
 \includegraphics[width=\linewidth]{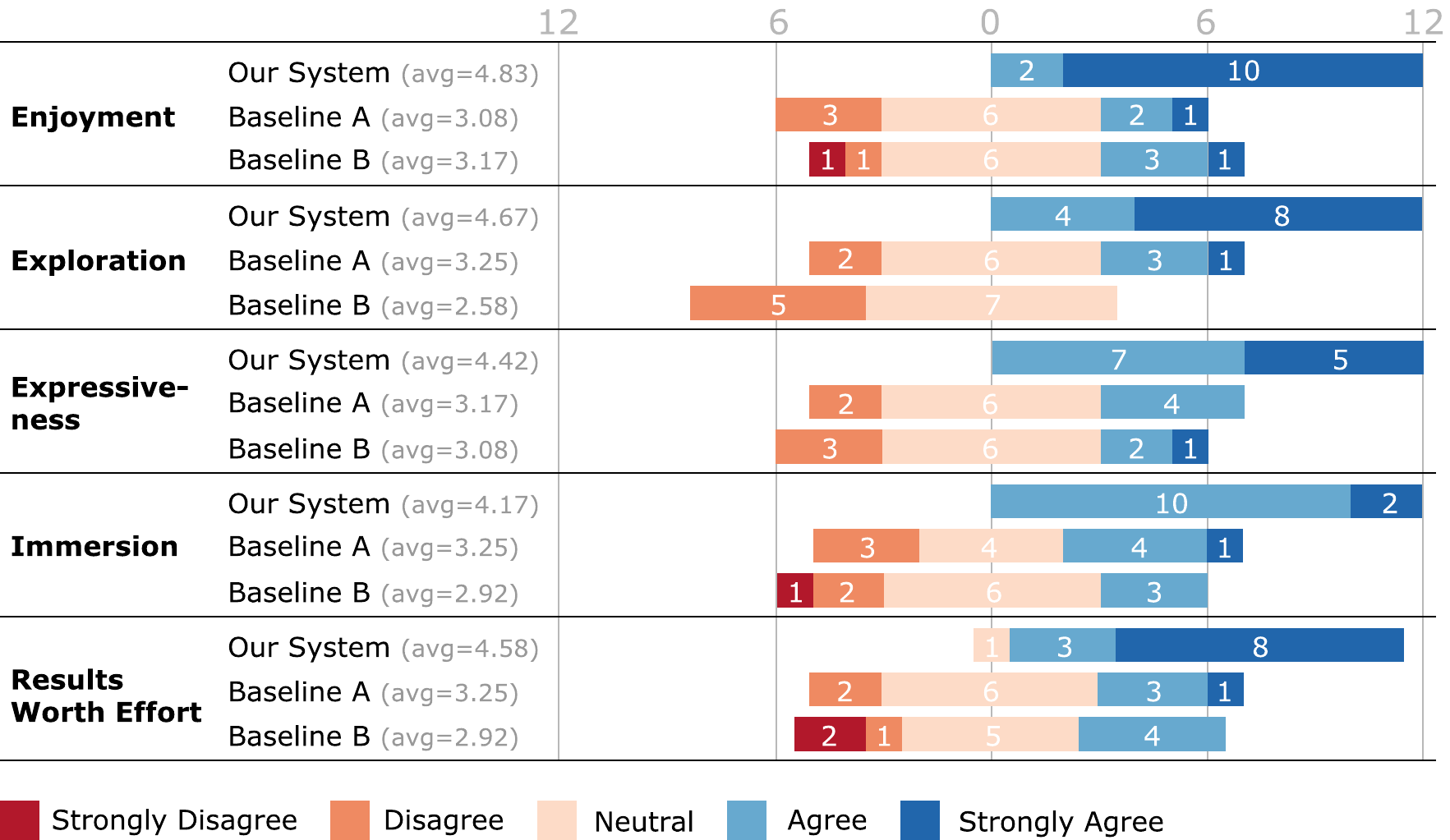}
 \caption{The results of the questionnaire regarding the creativity support of our system and two baseline systems.}
 \label{fig:comparison}
\end{figure}

\subsubsection{Creativity Support Comparison}
To compare our system with two baselines in facilitating creation tasks, we asked the participants to rate three systems based on the Creativity Support Index \cite{cherry2014quantifying}.
We excluded the inapplicable ``collaboration'' dimension.
The results are shown in \autoref{fig:comparison}.
Overall, our system outperforms the baselines in all dimensions, suggesting higher effectiveness in facilitating text-to-image creation.
We have summarized user feedback on their experience as follows:

\textbf{Our system vs. Baseline A.}
When browsing the images in Baseline A, the participants always had to review the original prompts one by one and compare them to find useful keywords.
This became more tedious when comparing multiple long prompt sentences.
\feng{In contrast, our system recommended important keywords for the image results and highlighted the keywords in the context, thus ``\textit{facilitating the selection and analysis of alternative keywords}'' (P11) and even ``\textit{motivating more creation attempts}'' (P5).
Moreover, some participants preferred to select multiple keywords and copy them all at once, which helped them improve input efficiency and focus more on the creation process.}

\textbf{Our system vs. Baseline B.}
\feng{When using Baseline B for prompt improvement, the participants found that it automatically added a set of keywords to the original prompts, mostly the disciplines of art and the names of the artists.
While this strategy was generally perceived as ``\textit{straightforward}'' (P2), it often ``\textit{led to the emergence of unexpected image styles}'' (P7) and sometimes even ``affected the accuracy of the image subjects'' (P3).
Notably, grappling with the suggested textual keywords in the absence of visual hints posed a considerable challenge.
Our system visualized the prompt keywords and images together, which ``\textit{facilitated users in comprehending the effect of the keywords}'' (P11) and even ``\textit{stimulated greater creativity in image generation}'' (P4).}

\subsection{Notable Observations}
\rw{
We have observed three prominent patterns in the way users craft their prompts.
The first and most common pattern involved users starting with basic sentences and progressively enhancing the image details by adding new keywords, as described in the usage scenarios.
The second pattern arose when users were unable to achieve desired results with detailed descriptions and needed summarizing suggestions.
\feng{For instance, using ``\textit{summer afternoon}'' to represent the coloring of the sky and clouds achieved the desired color effect and avoided color confusion with other objects.}
The third type involved the selective replacement of certain keywords.
For instance, the prompt ``\textit{a woman in an Arabesque costume lying on a soft Persian carpet}'' generated an image of a woman standing against a carpet backdrop. 
However, replacing ``carpet'' with ``pillow'' improved both body posture and background.
These prompting patterns and user intents can be detected for adaptive recommendations to select and refine keywords.
}

%% file: chapters/8-discussion.tex
% \newpage
\section{Discussion}
\rc{Apart from the case studies and user study, we conducted interviews with five experts from the Midjourney creator community to evaluate {\name}.
Their expertise spans from professional creators to advertisement designers, providing a diverse spectrum of insights.
Each interview session consists of case study walkthroughs, open-ended exploration, and semi-structured interviews.
\xingbo{Here, we distill design implications and discuss our system's generalizability and limitations.}
}

\subsection{Design Implications}
\rc{\textbf{Ideating with text-to-image creation.}
{\name} employs a breadth-first search approach to navigate the vast artistic search space.
This approach's efficiency surpasses traditional manual drafting methods, making it particularly beneficial for those more imaginatively inclined than technically skilled.
By turning abstract ideas into concrete visuals, it brings early-stage clarity to synchronize ideas across collaborators and simplify subsequent fine-tuning tasks, thus streamlining the overall creative process.
Unlike conventional image browsing tools, {\name} combines querying image databases with generating ad-hoc images, harnessing the ever-evolving capabilities of generative AI.
However, a knowledge gap exists for both novices and experts in adapting their prompt strategies alongside the model improvements, primarily because most users lack systematic exploratory methods for effective prompting.
Our approach to prompt keyword discovery is viable for exploring the expanding capabilities of text-to-image models.}

\rc{\textbf{Balancing between the brevity and effectiveness of prompts.}
Text-to-image creation has shifted the paradigm from the mechanical task of drawing to the perceptual notion of aesthetics. 
While drawing more details usually enhances an artwork, our findings indicate that for crafting prompts, less is often more.
Short, concept-focused prompts tend to yield better results than long, descriptive ones.
For instance, adding negative prompts to rectify a three-handed human portrait might unintentionally result in a handless figure because of the linguistic ambiguity and unpredictable control of prompt keywords over specific visual elements.
This phenomenon, which we term ``\textit{semantic contamination},'' discourages the unnecessary inclusion of ambiguous keywords but encourages the appropriate selection of effective keyword combinations.
This suggests the necessity for further research to identify the optimal balance between brevity and effectiveness of the prompt keywords to achieve the desired outcomes.}

\rc{\textbf{Incorporating aesthetic aspects into machine evaluation.}
Experts appreciate the efficiency that {\name} introduces to the artistic creation process, as it supports machine evaluation of massive generation and database recommendation. 
This evaluation can extend beyond the CLIP model to include existing art-based evaluators and integrate multiple concepts to reflect the multi-dimensional nature of aesthetic values.
For instance, detecting the ``uncanny valley''~\cite{uncanny_valley} requires the evaluation of human-like objects from several perspectives (\eg facial resemblance and limb authenticity), which opposing keywords alone may struggle to capture.
Future research could explore other evaluation metrics (\eg color analysis~\cite{feng_ipoet_2022} and art appreciation theories~\cite{zhang2023tcp}) to enhance artistic sensibilities within image evaluation.}

\rc{\textbf{Tracking the iterative creation process.}
Our interactive prompt engineering workflow supports users in articulating their requirements in the creation and bridging knowledge gaps in describing visual elements and artistic styles.
However, users often struggle with when to terminate the fine-tuning process in the user study due to uncertainty about the model's capability or inability to achieve the desired results~\cite{wang2023vis}. 
Currently, experts rely on their experience and heuristics to address the issue, highlighting a need for guidance in the iterative process.}

\rc{\textbf{Fine-tuning with multimodal interactions.}
AI-generated artworks are not bounded by physical laws or artistic principles, which often require fine-tuning of the generative output.
Image conditioning methods~\cite{rombach2022high, ramesh2022hierarchical, nichol2021glide} support text-guided inpainting to manipulate specific areas within images.
Another thread of research, exemplified by ControlNet~\cite{zhang2023adding} and DragGan~\cite{pan2023draggan}, provides interactive modifications (\eg specifying object boundaries and layouts), facilitating an iterative improvement process in text-to-image creation.
Experts express keen interest in these intuitive interactive tools that blend seamlessly into their existing workflows.
The combination of these research directions could lead to a more flexible and efficient creative process, especially if visualization tools can support high-level stylistic changes with textual prompts and nuanced detail adjustments through interactive inputs.}

\subsection{System Generalizability}
\xingbo{Besides facilitating prompt engineering for text-to-image generation, our system can be generalized to various applications, such as content moderation and model analysis.
For content moderation, our system can help examine and mitigate the potential misuse of text-to-image models. 
The prompt keyword recommendation and \textit{Local Explanation View} empower content reviewers to effectively identify and scrutinize harmful content prompted by specific groups of words and limit the usage of such prompts. 
For model analysis, the \textit{Image Browser View} allows researchers to evaluate a large number of generated model outputs. This feature facilitates the assessment of key aspects such as precision, stylization, and diversity in the model's output, thereby offering a comprehensive understanding of the model's strengths and weaknesses.}

\subsection{Limitations and Future Work}
% \textbf{Support more model hyper-parameters.}
{\name} currently allows users to configure text prompts and guidance scales.
Other hyper-parameters (\eg the \textit{inference step} for image denoising) can improve image quality but also pose challenges to the trial-and-error process.
We plan to support tuning multiple hyper-parameters to achieve flexible human control without increasing the learning cost or usage complexity.

% \textbf{Apply deep learning models in human-in-the-loop prompt engineering.}
{\name} retrieves similar images from DiffusionDB and identifies prompt keywords related to user prompts.
\feng{An interesting future work is to apply deep learning models, especially large language models, for prompt improvement.
For example, GPT-4~\cite{openai2023gpt} has been empowered with multi-modal data processing capability. 
It is possible to use GPT-4 to recommend prompt keywords for the image clusters or automatically revise user prompts according to the images of interest to the users.}
% Recent works in cross-modal representative learning (\eg CLIP model) have achieved significant progress to capture and align the semantics of images and text.
Integrating such models facilitates human-in-the-loop prompt improvement, leading to effective text-to-image generation.

%% file: chapters/9-conclusion.tex
\section{Conclusion}
This paper presents {\name}, a visual analytic system for interactive prompt engineering in text-to-image creation.
The system helps the users generate and explore a collection of image results and iteratively refine the input prompt.
We design a prompt recommendation model to recommend special and related prompt keywords for user prompts, which are mined from DiffusionDB with semantic-based retrieval and hierarchical keyword extraction.
Both the prompt keywords and their corresponding images are co-embedded in 2D space to facilitate interactive exploration and support personalized evaluation.
We present two usage scenarios of our system and conduct a user study and expert interviews to validate its effectiveness and usability.
The results not only show that {\name} recommends useful prompt keywords and facilitates interactive exploration, but also provide new insights for designing and improving prompting methods and the visual system for text-to-image creation.